\documentclass[lettersize,journal]{IEEEtran}
\usepackage{amsfonts, amsmath}
\usepackage{url}
\usepackage{graphicx}
\usepackage{cite}
\usepackage{mathrsfs}
\usepackage{ulem}
\usepackage{bbm}
\usepackage{makecell}
\usepackage{booktabs}
\usepackage{pifont}
\usepackage[table]{xcolor}
\usepackage[switch]{lineno}
\usepackage[utf8]{inputenc}
\usepackage[T1]{fontenc}
\usepackage[colorlinks=true, urlcolor=blue, linkcolor=red]{hyperref}
\newcolumntype{L}[1]{>{\raggedright\arraybackslash}p{#1}}
\newcolumntype{C}[1]{>{\centering\arraybackslash}p{#1}}
\newcolumntype{M}[1]{>{\centering\arraybackslash}m{#1}}
\newcolumntype{R}[1]{>{\raggedleft\arraybackslash}p{#1}}
\usepackage{scalerel}
\usepackage{tikz}
\usetikzlibrary{svg.path}
\newcommand{\settablefont}{\fontsize{6.5}{9.5}\selectfont}
\definecolor{orcidlogocol}{HTML}{A6CE39}
\tikzset{
  orcidlogo/.pic={
    \fill[orcidlogocol] svg{M256,128c0,70.7-57.3,128-128,128C57.3,256,0,198.7,0,128C0,57.3,57.3,0,128,0C198.7,0,256,57.3,256,128z};
    \fill[white] svg{M86.3,186.2H70.9V79.1h15.4v48.4V186.2z}
                 svg{M108.9,79.1h41.6c39.6,0,57,28.3,57,53.6c0,27.5-21.5,53.6-56.8,53.6h-41.8V79.1z M124.3,172.4h24.5c34.9,0,42.9-26.5,42.9-39.7c0-21.5-13.7-39.7-43.7-39.7h-23.7V172.4z}
                 svg{M88.7,56.8c0,5.5-4.5,10.1-10.1,10.1c-5.6,0-10.1-4.6-10.1-10.1c0-5.6,4.5-10.1,10.1-10.1C84.2,46.7,88.7,51.3,88.7,56.8z};
  }
}

\newcommand\orcidicon[1]{\href{https://orcid.org/#1}{\mbox{\scalerel*{
\begin{tikzpicture}[yscale=-1,transform shape]
\pic{orcidlogo};
\end{tikzpicture}
}{|}}}}

\usepackage{hyperref} 

\usepackage{multirow}
\begin{document}
\title{DepthMatch: Semi-Supervised RGB-D Scene Parsing through Depth-Guided Regularization}

\author{Jianxin Huang$^{\orcidicon{0009-0004-6558-9589}}$,~\IEEEmembership{Student Member,~IEEE}, Jiahang Li$^{\orcidicon{0009-0005-8379-249X}}$,~\IEEEmembership{Graduate Student Member,~IEEE}, \\
Sergey Vityazev,~\IEEEmembership{Senior Member,~IEEE}, Alexander Dvorkovich$^{\orcidicon{0000-0003-1190-3582}\,}$, and Rui Fan$^{\orcidicon{0000-0003-2593-6596}}$,~\IEEEmembership{Senior Member,~IEEE}

\thanks{
This work was supported in part by the National Natural Science Foundation of China under Grant 62473288 and Grant 62233013; in part by the Fundamental Research Funds for the Central Universities, NIO University Program (NIO UP); in part by the Xiaomi Young Talents Program. (\textit{Corresponding author: Rui Fan}).
}
\thanks{{Jianxin Huang, Jiahang Li, and Rui Fan are with the College of Electronics \& Information Engineering, Shanghai Institute of Intelligent Science and Technology, Shanghai Research Institute for Intelligent Autonomous Systems, the State Key Laboratory of Intelligent Autonomous Systems, and Frontiers Science Center for Intelligent Autonomous Systems, Tongji University, Shanghai 201804, China (e-mails: \{jasonhuang, lijiahang617\}@tongji.edu.cn, {rui.fan@ieee.org}).}}
\thanks{Sergey Vityazev is with Ryazan State Radio Engineering University, Ryazan 390035, Russian Federation (e-mail: vityazev.s.v@ieee.org).}
\thanks{Alexander Dvorkovich is with the Multimedia Technology and Telecom Department, Telecommunications Center, Moscow Institute of Physics and Technology, 141701, Institutsky Lane, 9, Dolgoprudny, Moscow Region, Russian Federation (e-mail: dvork.alex@gmail.com).}

\thanks{$^{}$
        {\tt\small }}
}

\markboth{IEEE SIGNAL PROCESSING LETTERS}{}
\maketitle
\begin{abstract}RGB-D scene parsing methods effectively capture both semantic and geometric features of the environment, demonstrating great potential under challenging conditions such as extreme weather and low lighting. However, existing RGB-D scene parsing methods predominantly rely on supervised training strategies, which require a large amount of manually annotated pixel-level labels that are both time-consuming and costly. To overcome these limitations, we introduce DepthMatch, a semi-supervised learning framework that is specifically designed for RGB-D scene parsing. To make full use of unlabeled data, we propose complementary patch mix-up augmentation to explore the latent relationships between texture and spatial features in RGB-D image pairs. We also design a lightweight spatial prior injector to replace traditional complex fusion modules, improving the efficiency of heterogeneous feature fusion. Furthermore, we introduce depth-guided boundary loss to enhance the model's boundary prediction capabilities. Experimental results demonstrate that DepthMatch exhibits high applicability in both indoor and outdoor scenes, achieving state-of-the-art results on the NYUv2 dataset and ranking first on the KITTI Semantics benchmark.  Our source code will be publicly available at \url{mias.group/DepthMatch}.
\end{abstract}


\section{INTRODUCTION}
\IEEEPARstart{S}{CENE} parsing task aims to assign each pixel in an image to a specific category and has broad applications in fields such as autonomous driving and robotics \cite{zhang2024lightweight, fan2020sneroadseg,wu2024s,feng2024sne}. With the widespread adoption of deep learning techniques, deep learning models trained via supervised learning have shown significant performance improvements in various image segmentation tasks compared to traditional geometry-based methods \cite{fan2019road, long2015fully, fan2019pothole}. However, single-modality networks relying solely on RGB images experience a substantial performance drop under challenging conditions, such as poor lighting or adverse weather \cite{huang2024roadformer, zhou2021mffenet, zhou2024mdnet}. To tackle this problem, FuseNet \cite{hazirbas2017fusenet} first explores the fusion of heterogeneous features derived from RGB and depth images, leveraging spatial information from depth data to improve scene parsing performance under extreme conditions. To fully exploit the complementary information in heterogeneous features, RoadFormer series \cite{huang2024roadformer,li2023roadformer} designs a novel attention-based feature fusion strategy, significantly enhancing the capability of feature integration. DPLNet \cite{dong2023efficient} introduces a lightweight multimodal prompt learning module to fuse RGB-D features, reducing training overhead while maintaining superior results.

Although depth information significantly improves scene parsing performance, existing \mbox{RGB-D} scene parsing methods still primarily rely on supervised training strategies to achieve satisfactory results, which require extensive pixel-level annotations. This limitation greatly hinders the deployment of advanced models, especially when sufficient annotations are unavailable \cite{wang2024depth}. Semi-supervised learning reduces annotation reliance by leveraging inexpensive unlabeled data to enhance scene parsing performance. Recent semi-supervised approaches commonly adopt pseudo-labeling frameworks based on consistency regularization \cite{yang2024unimatch}, where the model is required to produce consistent predictions under different augmentations or perturbations applied to the unlabeled data, thus improving the model's applicability across diverse conditions. In this context, CowMix \cite{french2019semi_iclr} demonstrates that combining Cutout \cite{devries2017improved} and CutMix \cite{yun2019cutmix} into unlabeled images is an effective regularization technique. UniMatch \cite{yang2023revisiting} inherits the weak-to-strong consistency regularization method from FixMatch \cite{sohn2020fixmatch}, constructing a simple yet effective regularization framework. Unfortunately, in the field of RGB-D scene parsing, semi-supervised training strategies remain underexplored.

To overcome the aforementioned challenges, we propose DepthMatch, a semi-supervised learning framework specifically designed for RGB-D scene parsing. Our method is built on consistency regularization, training the model to generate consistent predictions on the same data under various strong augmentations to fully leverage unlabeled RGB-D image pairs. Specifically, we design Complementary Patch Mix-up Augmentation strategy (CPMA) to randomly replace the color information with geometric information, encouraging the model to further explore the texture and spatial features in the input data. Considering that RGB images usually contribute more than depth images in RGB-D scene parsing tasks, we propose a Lightweight Spatial Prior Injector (LSPI) to replace traditional complex fusion modules, achieving efficient integration of heterogeneous features. Additionally, we introduce the Depth-Guided Boundary Loss (DGBL) to guide the model's boundary learning on unlabeled data. Experimental results demonstrate that DepthMatch, incorporating all these innovative components and strategies, achieves state-of-the-art (SoTA) performance on the NYUv2 and KITTI datasets, validating its strong advantages and high applicability across both indoor and outdoor scenes.

Our contributions can be summarized as follows:

\begin{itemize}

\item We introduce CPMA, a regularization strategy specifically designed for RGB-D scene parsing tasks, which encourages the model to better explore the relationships between texture and spatial features.

\item We propose LSPI to achieve more efficient feature fusion. It efficiently injects spatial information from depth images into RGB features during the encoding phase.

\item We propose DGBL, which imposes boundary supervision on unlabeled data based on contour information extracted from depth images.

\end{itemize}

\begin{figure}[t!]
\includegraphics[width=0.49\textwidth]{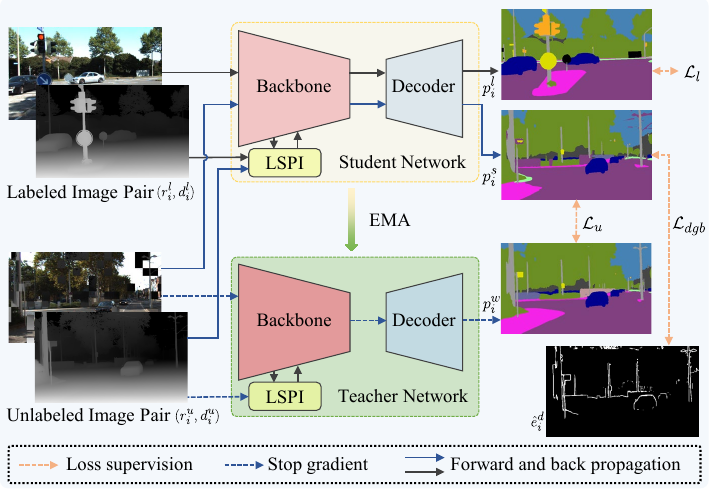}
\caption{An overview of our proposed DepthMatch.}
\label{fig.structure}
\end{figure}

\section{METHODOLOGY}
\label{Sect.methodology}

\subsection{Overall Training Pipeline}

The overall architecture of our proposed DepthMatch is illustrated in Fig. \ref{fig.structure}. Following \cite{yang2024unimatch}, we utilize the pre-trained DINOv2 \cite{oquab2023dinov2} model as our encoder, leveraging its self-supervised contrastive learning approach to enhance the quality and generalization of the extracted features. Meanwhile, we employ the Transformer-based DPT \cite{ranftl2021vision} as the decoder to generate globally consistent predictions. Unlike traditional supervised RGB-D scene parsing methods \cite{zhang2024lightweight, xiang2024self, yin2024dformer}, DepthMatch simultaneously utilizes a labeled dataset $D_l=\{(r_i^l, d_i^l, m_i^l)\}$ and a larger unlabeled dataset $D_u=\{ (r_i^u, d_i^u) \}$ for training, where $r_i$ and $d_i$ represent the $i$-th RGB-D image pair, and $m_i$ is the corresponding semantic ground truth mask. Following \cite{tarvainen2017mean}, we employ an EMA teacher with the same architecture to generate stable and improved pseudo-labels for unlabeled data. During training, each batch for both the teacher and student networks consists of $B_l$ labeled image pairs and $B_u$ unlabeled image pairs. For labeled data, we use manually annotated labels to supervise model training. The supervised loss $\mathcal{L}_{l}$ for labeled data is defined as:

\begin{equation}
\mathcal{L}_{l} = \frac{1}{B_l} \sum_{i=1}^{B_l} \mathcal{L}_{cls}(p_i^l, m_i^l),
\end{equation}
where $p_i^l$ represents the model's prediction for the $i$-th labeled image pair, $m_i^l$ is the corresponding ground truth mask, and $\mathcal{L}_{cls}$ is the pixel-wise cross-entropy loss for class prediction. For unlabeled image pairs, the model first generates pseudo-labels $p_i^w$ on weakly augmented image pairs $(r_i^{w}, d_i^{w})$, which are then used to supervise the learning of their strongly augmented counterparts $(r_i^{s}, d_i^{s})$. During semi-supervised training, weak augmentations include random scaling and random flipping of RGB-D image pairs, while strong augmentations encompass our proposed CPMA as well as color jittering, grayscaling, and Gaussian blurring. By training the model to make invariant predictions under various data augmentations, it can learn and extract latent patterns and features from unlabeled images \cite{sohn2020fixmatch}. The unsupervised loss $\mathcal{L}_{u}$ thus can be formulated as:

\begin{equation}
\mathcal{L}_{u} = \frac{1}{B_u} \sum_{i=1}^{B_u} \mathbbm{1}(\max(p_i^w) \geq \tau) \mathcal{L}_{cls}(p_i^s, \hat{p}_i^w),
\end{equation}
where $\hat{p}_i^w$ is the pseudo-label derived from $p_i^w$ using the argmax operation, $p_i^s$ represents the model's prediction for $(r_i^{s}, d_i^{s})$, and $\mathbbm{1}(\max(p_i^w) \geq \tau)$ introduces a predefined confidence threshold $\tau = 0.95$ to reduce noise in pseudo-labels by excluding those that do not meet the threshold from training \cite{yang2024unimatch}. This ensures that during the early training iterations, the model primarily learns from high-quality manually labeled images, gradually expanding to high-confidence pseudo-labeled images as training progresses \cite{sohn2020fixmatch}.

\subsection{Complementary Patch Mix-up Augmentation}

The consistency regularization-based semi-supervised algorithms require the model to make consistent predictions on unlabeled images under different data augmentations \cite{sohn2020fixmatch, yang2023revisiting}. However, simply transplanting strong data augmentation methods designed for RGB images to RGB-D data often leads to suboptimal results, as it overlooks the inherent modality differences between RGB-D image pairs \cite{wang2024depth}. To address this issue, we design a simple yet effective CPMA based on the modality-specific characteristics of RGB-D data. Previous studies have demonstrated that RGB images contain rich color and texture information, while depth images provide accurate geometric cues that are complementary in terms of detail and structure \cite{zhou2022frnet, zhou2025feature}. By swapping certain patches between the RGB image and its corresponding depth image during strong augmentation, CPMA replaces the original color and texture information of the patches with geometric information. This encourages the model to better explore the spatial features within the input data, resulting in improved neural representations \cite{gong2023continuous}. Specifically, CPMA can be formulated as follows:

\begin{equation}
\hat{r}_{i}^u = M \cdot r_{i}^u + \hat{M} \cdot d_{i}^u ,
\end{equation}
where $M$ and $\hat{M}=1-M$ form a pair of complementary random patch masks. Following \cite{shin2024complementary}, we set the patch size of the mask to 32 pixels, and the masking ratio for the RGB image will be further discussed in Sect \ref{Sect.ablation}.

\subsection{Lightweight Spatial Prior Injector}
Feature fusion modules that enhance and integrate RGB and depth features are crucial in RGB-D scene parsing networks \cite{valada2020self}. However, using complex feature fusion modules to symmetrically integrate heterogeneous features often introduces unnecessary overhead parameters \cite{dong2023efficient}. Inspired by \cite{dong2023efficient}, we propose an LSPI to progressively inject spatial priors from the depth image into RGB features during the encoding process, achieving effective fusion of heterogeneous features. Specifically, the RGB-D image pair $(r_i, d_i)$ is encoded into $(\boldsymbol{F}^r_i, \boldsymbol{F}^d_i)$ using two patch embedding layers. The heterogeneous embeddings $(\boldsymbol{F}^r_i, \boldsymbol{F}^d_i)$ are then downsampled to inject depth priors into the RGB embeddings, and the fused embeddings are then upsampled to their original dimensions. The LSPI can be formulated as:

\begin{equation}
\hat{\boldsymbol{F}^r_i} = w_f \big( w_r(\boldsymbol{F}^r_i) + w_d(\boldsymbol{F}^d_i) \big) + \boldsymbol{F}^r_i,
\end{equation}
where the linear layers $w_r$ and $w_d$ reduce the feature dimensions of $\boldsymbol{F}^r_i$ and $\boldsymbol{F}^d_i$ to one-quarter of their original size, and the linear layer $w_f$ fuses these heterogeneous embeddings and restores them to their original dimensions. The enhanced embedding $\hat{\boldsymbol{F}^r_i}$ is then processed through the encoder and decoder to produce the final predictions.

\subsection{Depth-Guided Boundary Loss}

Recent studies \cite{wang2022active} have demonstrated that introducing boundary supervision can improve model performance in handling complex object boundaries. This improvement arises because cross-entropy loss independently computes the classification accuracy of each pixel, lacking direct guidance for locating and precisely segmenting boundary pixels \cite{wu2023conditional}. In contrast, boundary supervision encourages the model to focus more on the distribution of boundary pixels \cite{wang2022active}. Unfortunately, existing boundary supervision methods \cite{wang2022active, wu2023conditional, qiu2024guided} rely on boundary information extracted from masks, making them unavailable for unlabeled images in semi-supervised learning. To overcome these limitations, we propose the DGBL to further guide the model in aligning predicted boundaries with ground-truth boundaries on unlabeled images. More concretely, we compute the boundary maps $e_i^p$ and $e_i^d$ from the model's predicted mask $p_i^u$ and the corresponding depth image $d_i^u$, respectively, and binarize the resulting boundary maps using a threshold of 0.1 to obtain $\hat{e}_i^p$ and $\hat{e}_i^d$. Finally, we use the mean squared error to measure the difference between the predicted and ground truth boundaries. The DGBL can be formulated as follows:

\begin{equation}
\mathcal{L}_{dgb} = \frac{1}{N} \sum_{j=1}^{N} y_{ij}^p \cdot (y_{ij}^p - y_{ij}^d)^2,
\end{equation}
where $y_{ij}^p$ and $y_{ij}^d$ represent the boundary values in $\hat{e}_i^p$ and $\hat{e}_i^d$, respectively, and $N$ is the total number of pixels in the image. By multiplying the loss for each pixel by $y_{ij}^p$, we ensure that the loss is only calculated for boundary pixels in the predicted mask map.

\subsection{Overall Loss Function}

At the beginning of training, the model often struggles to generate pseudo labels with well-defined boundaries \cite{yang2023revisiting}. As training progresses, the quality of these pseudo labels improves. Therefore, we gradually decrease the weight of $\mathcal{L}_{\text{dgb}}$ during training to increase the model's focus on $\mathcal{L}_u$ for unlabeled data. Inspired by previous studies \cite{yang2023effective}, we introduce a weight decay strategy to dynamically adjust the weight of $\mathcal{L}_{\text{dgb}}$. The time-dependent function is defined as $ f(t) = 1 - \frac{t - 1}{t_{\text{max}}}$, where $t = (1, 2, \dots, t_{\text{max}})$ represents the current training epoch, and $t_{\text{max}}$ denotes the maximum training epoch. For each batch of training data, we minimize the following loss function:

\begin{equation}
\mathcal{L}_{total} = \mathcal{L}_{l} + \lambda_u \big( \mathcal{L}_{u} + f(t) \mathcal{L}_{dgb} \big),
\end{equation}
where $\lambda_u$ balances the influence of the unlabeled data, while $f(t)$ controls the weight of $\mathcal{L}_{dgb}$ and progressively decays during training.

\section{EXPERIMENTS}
\label{Sect.experiments}

\subsection{Datasets and Evaluation Metrics}
\label{Sect.dataset}

We conduct experiments on RGB-D scene parsing datasets for both indoor and outdoor environments. For indoor scenes, we used the popular NYUv2 dataset, which consists of 1,449 labeled RGB-D image pairs. Following the standard partition, the dataset was divided into 795 training samples and 654 testing samples. All images are resized to a resolution of 630$\times$476 pixels. On the other hand, we evaluate the performance of DepthMatch in outdoor scenarios using the KITTI Semantics \cite{abu2018augmented} dataset. This dataset contains 200 labeled images across 19 categories and an additional 4,000 unlabeled images. All images are resized to a resolution of 1,274$\times$378 pixels, with depth data obtained using ViTAStereo \cite{liu2024playing}. We evaluate model performance using the commonly adopted metrics: mean Intersection over Union (mIoU) and instance Intersection over Union (iIoU). We also use model parameters (Params) and floating-point operations (FLOPs) to assess the model efficiency. Additionally, the evaluation metrics used for the KITTI Semantics benchmarks are available on the official webpage: \url{www.cvlibs.net/datasets/kitti/eval_semseg.php?benchmark=semantics2015}.

\subsection{Implementation Details}
\label{Sect.setup}

Our network is trained using the AdamW optimizer \cite{loshchilov2018decoupled}, with a polynomial decay strategy for the learning rate \cite{chen2017deeplab}. The initial learning rate is set to $2 \times 10^{-4}$ with a weight decay of $10^{-2}$, and learning rate multipliers of $2.5 \times 10^{-2}$ are applied to the weight-sharing backbone. It is worth noting that for the KITTI Semantics, we employed a multi-scale inference strategy with scaling factors of $\{1.05, 1.5, 2.0, 2.5\}$.

\begin{table}[t!]
\settablefont
\centering
\caption{Comparison of DepthMatch's performance with SoTA algorithms on the test set of the NYUv2 dataset. }

\setlength{\tabcolsep}{2.2pt}
\begin{tabular}{l|cc|ccc}
\toprule[1pt]
Method &  Publication &  Backbone & Params (M) $\downarrow$ & FLOPs (G) $\downarrow$ & mIoU (\%) $\uparrow$ \\ 
\hline
ACNet & ICIP'19 \cite{hu2019acnet} & ResNet-50  & 116.6 & 126.7 & 48.1  \\
FRNet & JSTSP'22 \cite{zhou2022frnet} & ResNet-34  & - & - & 53.6  \\
CMX & T-ITS'23 \cite{zhang2023cmx} & MiT-B5  & 181.1 & 167.8 & 56.9  \\
Sigma & WACV'25 \cite{wan2024sigma} & VMamba-S  & 69.8  & 138.9 & 57.0  \\
DFormer & ICLR'24 \cite{yin2024dformer} & DFormer-L  & 39.0 & 65.7 & 57.2  \\
RoadFormer+ & T-IV'24 \cite{huang2024roadformer} & ConvNeXt-B  & 152.4 & 281.5 & 57.6  \\
FCDENet & IOTJ'25 \cite{zhou2025feature} & MiT-B4  & 128.9 & - & 57.6  \\
GeminiFusion & ICML'24 \cite{jiageminifusion} & MiT-B5  & 137.2 & 300.5  & 57.7  \\
SEFNet & SPL'24 \cite{xiang2024self} & ResNet-101 & 90.7 & 118.1 & 58.5  \\
DPLNet & IROS'24 \cite{dong2023efficient} & MiT-B5  & 88.6 & - & 59.3  \\

\hline
\rowcolor{gray!20} \textbf{DepthMatch (1/4)}  & Ours & DINOv2-S & {26.9} & 68.1 & 56.6 \\
\rowcolor{gray!20} \textbf{DepthMatch (1/2)}  & Ours & DINOv2-S & {26.9} & 68.1 & 58.7 \\
\rowcolor{gray!20} \textbf{DepthMatch}  & Ours & DINOv2-S & {26.9} & 68.1 & 59.9 \\
\rowcolor{gray!20} \textbf{DepthMatch*}  & Ours & DINOv2-S & {26.9} & 68.1 & \textbf{61.4} \\
\bottomrule[1pt]
\end{tabular}
\\[2pt] 
\footnotesize{* Using SUN-RGBD \cite{song2015sun} training set as unlabeled data.}

\label{tab.nyuv2}
\end{table}

\begin{table}[t!]
\settablefont
\centering
\caption{Comparison on the KITTI Semantics benchmark in terms of mIoU and iIoU for both Class and Category.}
\setlength{\tabcolsep}{6pt}
\label{tab:comparison}
\begin{tabular}{lccccc}
\toprule[1pt]
{Method} & \multicolumn{2}{c}{{Class}} & \multicolumn{2}{c}{{Category}} & Rank\\
\cmidrule(lr){2-3} \cmidrule(lr){4-5}
 & {mIoU (\%)} & {iIoU (\%)} & {mIoU (\%)} & {iIoU (\%)} \\
\midrule
VideoProp \cite{zhu2019improving} & 72.82 & 48.68 & 88.99 & \textbf{75.26} & 5 \\
RoadFormer+ \cite{huang2024roadformer} & 73.13 & 45.88 & 88.75 & 73.46 & 4 \\
UJS \cite{cai2021multi} & 75.11 & 47.71 & 89.53 & 75.75 & 3 \\
WRP \cite{ganeshan2021warp} & 76.44 & \textbf{50.92} & 89.63 & 73.69 & 2 \\
\hline
\rowcolor{gray!20} \textbf{DepthMatch} & \textbf{76.60} & 48.48 & \textbf{90.16} & 75.05 & \textbf{1} \\
\bottomrule[1pt]
\end{tabular}
\label{tab.kittisemantics}
\end{table}

\begin{table}[t!]
\settablefont
\centering
\caption{Ablation study of masking ratio for CPMA}
\setlength{\tabcolsep}{12pt} 
\begin{tabular}
{c|ccccc}
\toprule[1pt]
{Masking Ratio} & {0} & {0.05} & {0.1} & {0.15} & {0.2}   \\
\hline
{mIoU (\%)} $\uparrow$  & 60.2 & 60.4 & \textbf{60.9} & 60.5 & 60.5   \\
\bottomrule[1pt]
\end{tabular}
\label{tab.Ablation_ratio}
\end{table}

\begin{table}[t!]
\settablefont
\centering
\caption{Ablation study of the module part of DepthMatch.}
\setlength{\tabcolsep}{6pt} 
\begin{tabular}
{c|ccc|cc}
\toprule[1pt]
Data Type & LSPI & CPMA & DGBL & Params (M) $\downarrow$& mIoU (\%) $\uparrow$   \\
\hline
RGB & $\times$& $\times$& $\times$& 24.9  & 59.6  \\
RGB-D  & $\checkmark$& $\times$& $\times$& 26.9  & 60.2   \\
RGB-D  & $\checkmark$& $\checkmark$& $\times$& 26.9  & 60.9  \\
RGB-D  & $\checkmark$& $\times$& $\checkmark$& 26.9  & 60.8  \\
\cline{1-6}
RGB-D  & $\checkmark$& $\checkmark$& $\checkmark$& 26.9 & \textbf{61.4}  \\
\bottomrule[1pt]
\end{tabular}
\label{tab.Ablation}
\end{table}

\subsection{Comparison with SoTA Networks}
\label{Sect.comparison}

\begin{figure}[t!]
\includegraphics[width=0.485\textwidth]{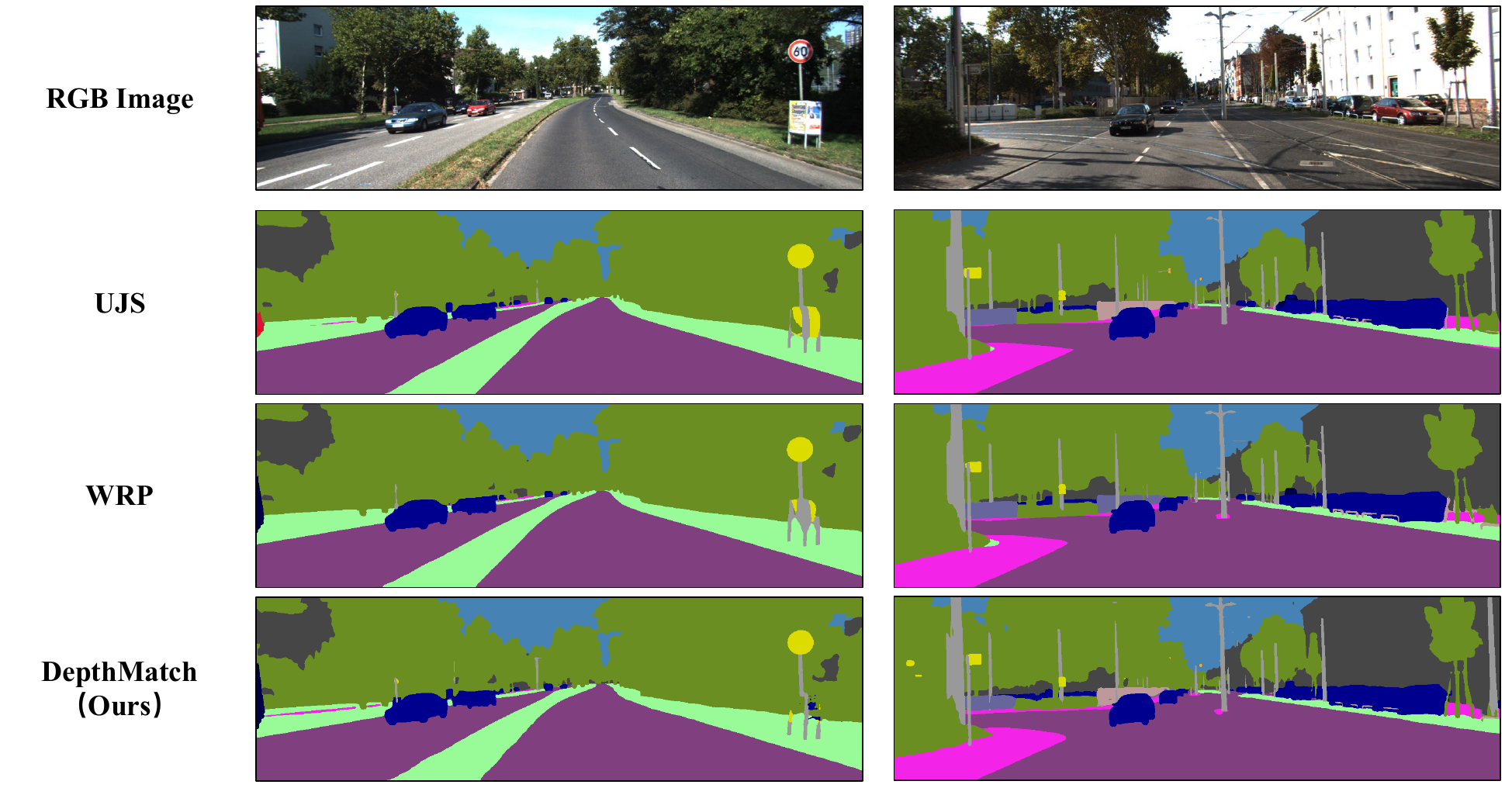}
\caption{Qualitative comparison on the KITTI Semantics dataset. The results are produced by the official KITTI online benchmark suite.}
\label{fig.visualizationkitti}
\end{figure}

To validate the effectiveness of our proposed DepthMatch, we compared it with multiple SoTA RGB-D scene parsing methods on two popular datasets, NYUv2 and KITTI Semantics. Quantitative experimental results demonstrate that DepthMatch significantly outperforms all other SoTA methods with fewer model parameters on the NYUv2 dataset. Even when trained with only 1/4 of the labeled images from the NYUv2 training set, DepthMatch maintains exceptional performance. Additionally, by incorporating supplementary unlabeled data, DepthMatch achieves 61.4\% mIoU, demonstrating its effective capability to leverage unlabeled data for enhanced model performance. Notably, DepthMatch yields an inference speed of 74 FPS when processing images with a resolution of 630$\times$476 pixels on an NVIDIA RTX 3090 GPU. To further verify the superiority of our method in outdoor driving scene parsing, we submitted the test results produced by \mbox{DepthMatch} to the KITTI Semantics benchmark for performance comparison. As shown in Table \ref{tab.kittisemantics}, \mbox{DepthMatch} ranks first on the KITTI Semantics benchmark. Fig. \ref{fig.visualizationkitti} further presents the qualitative comparison on the KITTI Semantics dataset, demonstrating the outstanding performance of our method in both global scene understanding and local boundary detection. These results fully indicate the robustness and high applicability of DepthMatch across indoor and outdoor scenes.

\subsection{Ablation Studies}
\label{Sect.ablation}

All ablation experiments are conducted on the NYUv2 dataset, using the SUN-RGBD training set as unlabeled data. We first investigate the optimal masking ratio for the CPMA. Based on the experimental results in Table \ref{tab.Ablation_ratio}, we set the masking ratio to 0.1. To further validate the effectiveness of our proposed DepthMatch, we perform ablation studies for each contribution, and all results are shown in Table \ref{tab.Ablation}. Specifically, we select UniMatchV2 \cite{yang2024unimatch} with a DINOv2-S  \cite{oquab2023dinov2} backbone as our baseline model and incrementally add the proposed LSPI, CPMA, and DGBL during training. The ablation results demonstrate that the LSPI module efficiently exploits spatial information from depth images, while the CPMA and DGBL enhance the training on unlabeled data, significantly boosting the model. These results fully validate the effectiveness of each part of the proposed DepthMatch.

\section{CONCLUSION}
\label{Sect.conclusion}

In this letter, we proposed DepthMatch, a lightweight yet powerful semi-supervised training framework specifically designed for RGB-D scene parsing, aimed at reducing the reliance on manually annotated data. Specifically, we employed the LSPI module to efficiently inject spatial priors from depth images into RGB features. Additionally, we designed CPMA and DGBL to provide more effective regularization strategies and boundary supervision for unlabeled image pairs. Extensive experiments demonstrated that DepthMatch effectively utilized inexpensive unlabeled data to improve model performance, surpassing existing SoTA methods. Our future work will focus on extending our framework to other scene parsing tasks.

\clearpage

\bibliographystyle{IEEEtran}
\bibliography{ref.bib} 


\end{document}